\documentclass[letterpaper, 10 pt, conference]{ieeeconf}  
\IEEEoverridecommandlockouts                              
\usepackage{graphicx}
\usepackage{amsmath,amssymb}

\usepackage{cite}
\usepackage{balance}
\usepackage{graphicx}
\usepackage{placeins}
\usepackage[ruled,vlined,linesnumbered]{algorithm2e}
\usepackage{algpseudocode}
\usepackage{enumerate}
\usepackage{multirow}
\usepackage{array}
\usepackage{booktabs}
\usepackage{threeparttable}
\usepackage{gensymb}
\usepackage{subfig}
\usepackage{xcolor}
\usepackage{comment}

\newcolumntype{L}[1]{>{\centering\arraybackslash}m{#1}}
\renewcommand{\tabcolsep}{3pt}

\newcommand{\ie}{{i.e.},\ }

\newcommand{\R}{{\mathbb{R}}}

\usepackage{caption}
\newtheorem{thm}{Theorem}[section]

\newtheorem{rem}[thm]{Remark}

\newtheorem{prob}[thm]{Problem}

\title{\LARGE \bf  Delayed Expansion AGT: Kinodynamic Planning with Application to Tractor-Trailer Parking}

\author{Dongliang Zheng, Yebin Wang, Stefano Di Cairano, and Panagiotis Tsiotras 
\thanks{D. Zheng and P. Tsiotras are with the School of Aerospace Engineering, Georgia Institute of Technology, Atlanta, GA 30332, USA. This work was done while D. Zheng was a research intern at Mitsubishi Electric Research Laboratories. (email: \tt\small \{dzheng,tsiotras\}@gatech.edu).}
\thanks{Y. Wang and S. Di Cairano are with Mitsubishi Electric Research Laboratories, Cambridge, MA 02139, USA (email: \tt\small \{yebinwang,dicairano\}@ieee.org).}
}

\begin{document}

\maketitle
\thispagestyle{empty}
\pagestyle{empty}

\begin{abstract}
Kinodynamic planning of articulated vehicles in cluttered environments faces additional challenges arising from high-dimensional state space and complex system dynamics. Built upon \cite{wang2019improved,LeuWanTom22}, this work proposes the DE-AGT algorithm that grows a tree using pre-computed motion primitives (MPs) and A* heuristics.
The first feature of DE-AGT is a delayed expansion of MPs. In particular, the MPs are divided into different modes, which are ranked online.
With the MP classification and prioritization, DE-AGT expands the most promising mode of MPs first, which eliminates unnecessary computation and finds solutions faster.
To obtain the cost-to-go heuristic for nonholonomic articulated vehicles, we rely on supervised learning and train neural networks for fast and accurate cost-to-go prediction.
The learned heuristic is used for online mode ranking and node selection. Another feature of DE-AGT is the improved goal-reaching.
Exactly reaching a goal state usually requires a constant connection checking with the goal by solving steering problems -- non-trivial and time-consuming for articulated vehicles. The proposed termination scheme overcomes this challenge by tightly integrating a light-weight trajectory tracking controller with the search process. DE-AGT is implemented for autonomous parking of a general car-like tractor with 3-trailer. Simulation results show an average of 10x acceleration compared to a previous method. 
\end{abstract}
\vspace{-1.5mm}

\maketitle
\thispagestyle{empty}
\pagestyle{empty}

\section{Introduction}\label{sec:intro}

High-dimensional state space, nonlinear dynamics, nonholonomic constraints, and cluttered environments pose great challenges for efficient kinodynamic planning despite recent advances. One example is the planning for tractor-trailer autonomous parking, where it may take several attempts for trained professionals to park into a confined space. Thus, efficient kinodynamic planning that can deal with complex problems such as tractor-trailer parking is desirable.

Existing kinodynamic planning methods may be divided into three classes: optimization-based methods, sampling-based methods, and state-lattice methods. Seeking a numerical solution of an optimal control problem (OCP), optimization-based methods entail a careful design of the representation of the general obstacle constraints and a good initial guess of the solution and may converge to local minima~\cite{Deits2015Efficient,Bonalli2019GuSTO,rao2009survey,zhang2018autonomous,bergman2018combining}.
For example, \cite{Deits2015Efficient,Deits2015Computing} first approximate the free space of the environment with connected convex hulls using a semi-definite program, then incorporate the obstacle constraints with integer variables, and formulate a mixed-integer program.
These approaches do not scale well with the number of obstacles. Sampling-based planners with simplified vehicle models have been used to provide an initial solution for the OCP \cite{lavalle2006planning, zhang2018autonomous,leu2021efficient, campos2017hybrid}. This treatment does not guarantee the feasibility of the OCP because the initial solution may not be homotopic to the true one. 

Sampling-based methods have shown promising results for planning in high-dimensional space. On one hand, they rely on sampling to explore collision-free space; on the other hand, they resort to either a steering function, e.g., RRT* and its variants \cite{Karaman2011Sampling,Karaman2010Optimal, Dustin2013Kinodynamic,Zheng2021Accelerating,WanJhaAke17}, or simulation\cite{LaValle2011Motion}, e.g., SST \cite{Li2016Asymptotically} to resolve dynamic feasibility. The steering function essentially solves a two-point boundary value problem (TPBVP). Algorithms such as Kino-RRT* \cite{Zheng2021Accelerating} and kinodynamic FMT* \cite{Schmerling2015Optimal} may be promising for simple (e.g., linear) systems where an analytic solution for the TPBVP is available. However, solving TPBVPs online for nonlinear systems for every possible edge extension is computationally prohibitive. Algorithms using a learning-based steering function are presented in \cite{Chiang2019RLRRT, zheng2021sampling}. SST avoids solving the TPBVP by using random control sampling and simulation. However, without the local optimal edges provided by the steering function, the convergence of SST to a good solution is slow.
In addition, due to the complexity of the tractor with 3-trailer system, generating reverse motion using random control input sampling will cause many invalid configurations, e.g. jack-knife.
Another challenge of sampling-based methods lies in the time-consuming goal reaching: exact reaching requires frequent checks of connection with the goal by solving steering problems and approximate reaching requires the arrival state is in a small neighborhood of the goal. 

State-lattice-based methods discretize state space into a pre-defined lattice \cite{Likhachev2009Planning, Pivtoraiko2009Differentially, cirillo2014lattice}, where neighboring nodes are connected using motion primitives (MPs). The MPs can be generated offline by solving OCPs to ensure dynamic feasibility. The lattice and MPs imply a graph for online search, e.g., A* \cite{HarNilRap68}, to find a plan. A combination of RRT* with MPs is studied in \cite{Sakcak2019Sampling} where samples are drawn from the lattice. State-lattice-based methods suffer from the curse of dimensionality and resolution-completeness~\cite{bergman2020improved}. Nevertheless, various planning works for tractor-trailer systems adopt the state-lattice framework \cite{ljungqvist2017lattice, ljungqvist2019path, tows2021reversing}, where dimensionality reduction is introduced. 

The state-lattice methods, referred to as on-grid methods below, unnecessarily restricts the vehicle to transit over the lattice. In our previous Work \cite{wang2019improved, LeuWanTom22}, an improved A-search guided tree (i-AGT) algorithm for passenger car and tractor-trailer systems is developed, where planning explores off-grid states and thus empirically results in better motion plans and faster computation. i-AGT extends A* by introducing mode-based node expansion where the MPs in the most promising mode are applied during node expansion. Real-world experiments that track the planned trajectory using a trajectory tracking controller are conducted in our recent paper \cite{You2023Integral}. The mode-based node expansion is analogous to partial expansion A* \cite{Yoshizumi2000A} and its enhanced version~\cite{Felner2012Partial}. 

\begin{figure}[t]
 \centering
   \begin{tabular}{@{}ccc@{}}
    \begin{minipage}{.15\textwidth}
    \includegraphics[keepaspectratio=true, width=1\linewidth]{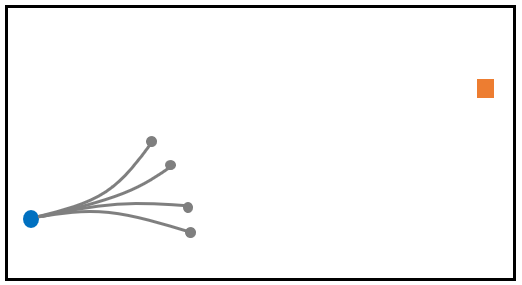}
    \captionof*{figure}{(a)} 
    \end{minipage} &
    \begin{minipage}{.15\textwidth}
    \includegraphics[keepaspectratio=true, width=1\linewidth]{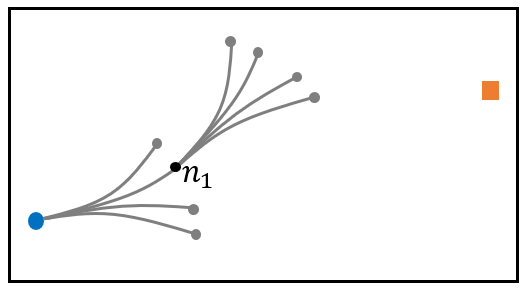}
     \captionof*{figure}{(b)}
   \end{minipage} &
   \begin{minipage}{.15\textwidth}
    \includegraphics[keepaspectratio=true, width=1\linewidth]{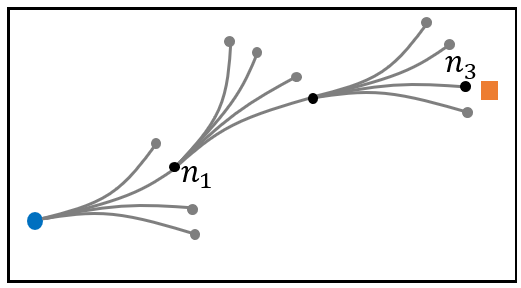}
    \captionof*{figure}{(c)}
   \end{minipage}
  \end{tabular}
  \begin{tabular}{@{}ccc@{}}
    \begin{minipage}{.15\textwidth}
    \includegraphics[keepaspectratio=true, width=1\linewidth]{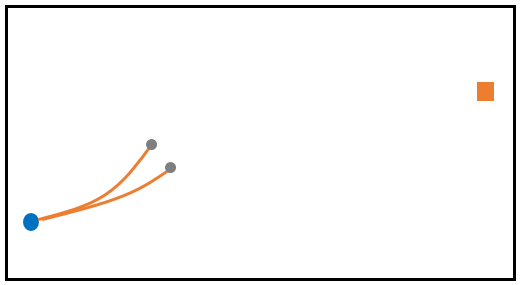}
    \captionof*{figure}{(d)} 
    \end{minipage} &
    \begin{minipage}{.15\textwidth}
    \includegraphics[keepaspectratio=true, width=1\linewidth]{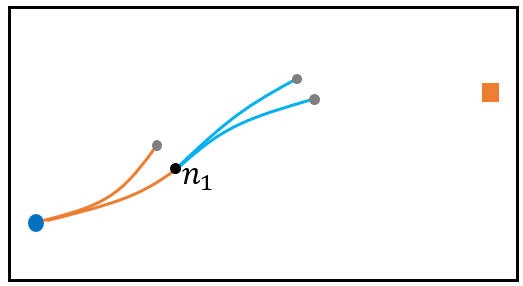}
     \captionof*{figure}{(e)}
   \end{minipage} &
   \begin{minipage}{.15\textwidth}
    \includegraphics[keepaspectratio=true, width=1\linewidth]{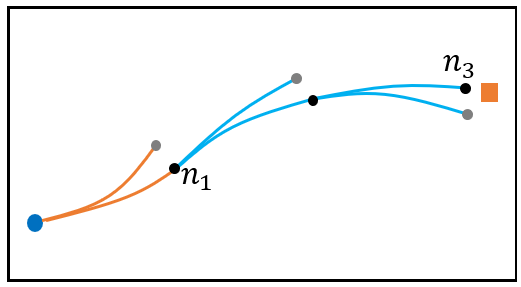}
    \captionof*{figure}{(f)}
   \end{minipage}
  \end{tabular}
    \caption{Motivation of delayed expansion method that prioritizes the most promising mode of MPs. A planning problem with 4 MPs is illustrated. (a)-(c) The AGT algorithm. After selecting a node, all 4 MPs are expanded. (d)-(f) DE-AGT algorithm. The MPs are divided into two modes, indicated by the colors orange and blue. After selecting a node, the two modes of MPs are ranked. Only the MPs in the best mode are expanded. Both algorithms find the same solution, while DE-AGT evaluates fewer MPs. A node may be selected multiple times in DE-AGT. For each time, the remaining best mode is expanded. A node is removed from the search queue when all modes are expanded.}
  \label{fig:DelayedExpansionMotivation}
\end{figure}

This paper presents the delayed expansion AGT (DE-AGT) algorithm, which is built on i-AGT. Major differences from \cite{wang2019improved,LeuWanTom22} are fourth-fold. First, DE-AGT adopts a new mode ranking method based on defining the cost-to-go of each mode, see Figure~\ref{fig:DelayedExpansionMotivation}. Second, DE-AGT features a tracking control-based termination scheme to improve goal reaching. Particularly, 
once a node enters a relatively large neighborhood of the goal, a trajectory tracking controller based on the linear quadratic regulator (LQR) is invoked to construct an LQR motion that may reach a target neighborhood of the goal. Third, DE-AGT exploits a neural network (NN) to approximate the cost-to-go of nodes and also uses the NN for online mode ranking.  The NN is trained by supervised learning. In contrast, \cite{LeuWanTom22} adopts NN and reinforcement learning to learn Q function which offers the basis for mode selection. Since the reverse dynamics of the tractor-trailer system is unstable and reverse motions rarely get reward, reinforcement learning ends up with a Q function heavily favoring forward motions and thus the resultant plan could be lengthy. Lastly, extensive simulations of a tractor-trailer system demonstrate its effectiveness. 

\section{Preliminaries}\label{sec:Preliminaries}

\subsection{Problem Statement}
Consider a system with the following dynamics
\begin{equation}\label{eq:plant}
\dot{X} = f(X,u),
\end{equation}
where $X \in \mathcal X \subset \R^{n_x}$ is the state, $u \in \mathcal U \subset \R^{n_u}$ is the control input, and $f$ is a smooth vector field. The free space is denoted by $\mathcal X_{\mathrm{free}} \subset \mathcal{X}$, where at $X \in \mathcal X_{\mathrm{free}}$, the system~\eqref{eq:plant} does not collide with any obstacles in the environment. 
The kinodynamic planning problem is as follows:

\begin{prob}\label{prob:path_planning}
Given $\mathcal X_{\mathrm{free}}$, an initial state $X_s \in \mathcal X_{\mathrm{free}}$, a goal state $X_g \in \mathcal X_{\mathrm{free}}$, system \eqref{eq:plant}, and a cost function $c(X, u)$, find the optimal trajectory $\mathcal P_t$ that:
\begin{enumerate}[(I)]
\item starts at $X_s$ and ends at $X_g$, while satisfying \eqref{eq:plant};
\item is collision-free $ \mathcal P_t \subset \mathcal X_{\mathrm{free}}$; and
\item minimizes the cost $\int_{0}^{t_{\mathrm{f}}} c(X,u) \, \mathrm{d}\tau$, where $t_{\mathrm{f}}$ is the final time. 
\end{enumerate}
\end{prob}

Problem \ref{prob:path_planning} can be formulated as the following OCP: 
\begin{equation}
\begin{split}
\min_{u, t_{\mathrm{f}}} \ \ & J = \int_{0}^{t_{\mathrm{f}}} c(X,u) \, \mathrm{d}\tau, \\
\mathrm{s.t.} \ \ & \dot{X} = f(X,u), \\
& X(0)=X_s, \ X(t_{\mathrm{f}})=X_g, \\
& u \in \mathcal{U}, \ X \in \mathcal{X}_{\mathrm{free}}, \ \forall t\in [0, t_{\mathrm{f}}].
\label{eq:OCP}
\end{split}
\end{equation}

We solve problem (\ref{eq:OCP}) approximately using MPs. Each MP is generated by solving an OCP similar to (\ref{eq:OCP}), with different boundary conditions and without obstacle constraints. We tackle Problem \ref{prob:path_planning} by repetitively applying MPs. The new planning problem is given below. 

\begin{prob}\label{prob:MP planning}
Given $\mathcal X_{\mathrm{free}}$, an initial state $X_s \in \mathcal X_{\mathrm{free}}$, a goal state $X_g \in \mathcal X_{\mathrm{free}}$, and a set of MPs, find the trajectory as the concatenation of a sequence of $n$ MPs, $\{mp_1, mp_2, \cdots, mp_n\}$, such that
\begin{enumerate}[(I)]
\item starts at $X_s$ and reaches a neighborhood of $X_g$;
\item is collision-free; and
\item minimizes the accumulated cost of $\{mp_1, \cdots, mp_n\}$. 
\end{enumerate}
\end{prob}

\subsection{Trailer Modeling}

\begin{figure}[t]
	\centering \vspace{-0.05cm}
    \includegraphics[keepaspectratio=true, width=0.7\linewidth]{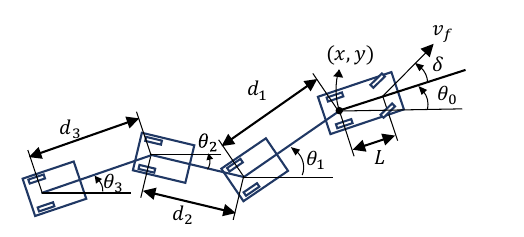}
	\caption{Kinematics of a front drive tractor with 3 trailers. All trailers are on-axle and all angles representing the orientation of the tractor and trailers ($\theta_i, i=0,\dots,3$) are relative to the $x$-axis.}
	\label{fig:trailer_kinematics_rough}
\end{figure}

Consider the front wheel drive \emph{standard tractor-trailer system} \cite{rouchon1993flatness,altafini2001feedback} shown in Figure \ref{fig:trailer_kinematics_rough}, where $(x,y)^\top$ are the coordinates of the midpoint of the tractor's rear wheel axis, $\theta_0$ is the tractor orientation, and $\theta_1, \theta_2, \theta_3$ are the orientations of trailers. The control inputs are $v_f$ and $\delta$, denoting the front-wheel velocity and the steering angle of the tractor, respectively. Mechanical constraint $|\delta| \le \delta_{\max}$ limits the minimum turning radius $R$.  Since $v_f$ and $\delta$ can be independently controlled, we introduce new control variables: $u = (v,s)^\top = (\cos(\delta) v_f, \tan(\delta)/\tan(\delta_{\max}))^\top$, 
where $\tan(\delta_{\max}) = L/R$, $L$ is the distance between $(x,y)$ and the midpoint of the front wheel axis, and $|s| \le 1$ is the normalized steering angle. Defining $\xi = (\xi_1, \xi_2, \xi_3, \xi_4, \xi_5, \xi_6)^\top = (x,y,\theta_0, \theta_1-\theta_0, \theta_2 - \theta_1, \theta_3 - \theta_2)^\top$, we have
\begin{equation}\label{eq:fwd_car_dynamics2}
	\begin{aligned}
		\dot x & = \cos(\theta_0)v \\
		\dot y & = \sin(\theta_0)v \\
		\dot \theta_0 & =  \frac{vs}{R} \\
		\dot \xi_4 & = -v\frac{d_1s +  \sin(\xi_4)R}{Rd_1} \\
		\dot \xi_5 & = -v \frac{d_1\cos(\xi_4)\sin(\xi_5) - d_2\sin(\xi_4)}{d_1d_2} \\
		\dot \xi_6 & = -v \cos(\xi_4) \frac{d_2\cos(\xi_5)\sin(\xi_6)-d_3\sin(\xi_5)}{d_2d_3}.
	\end{aligned}
\end{equation}
In $\xi$-coordinates, the constraints to avoid jack-knife are
\begin{equation}\label{eq:constraint_jackknife}
    |\xi_k| \le \xi_{\max}, \quad 4 \le k \le 6,
\end{equation}
where $\xi_{\max}$ must be less than $\pi/2$. The trailer system is subject to additional state and control constraints:
\begin{equation}\label{eq:constraint_xu}
    -\pi < \theta_0 \le \pi, \quad |v| \le v_{\max}, \quad |s| \le 1.
\end{equation}

\begin{rem}
    Throughout this paper, trailer system \eqref{eq:fwd_car_dynamics2} is used as an example to describe the proposed algorithm. The work can be generalized to other systems.
\end{rem}

\subsection{Circular equilibrium configurations} \label{subsec:Circular}
Searching over a 6D state-space system \eqref{eq:fwd_car_dynamics2} suffers from the curse of dimensionality. The idea of circular equilibrium configurations is borrowed to reduce the dimension of search space \cite{altafini2001feedback, ljungqvist2019path}.
We restrict $\xi$ to ensure that the tractor and trailers move in circles, hence trailers and tractor have the same yaw rate $C$. Given $v$, the yaw rate of the tractor and the headings of trailers are uniquely determined by the steering action $s$. Specifically, $\dot \theta_0 = vs/R = C(s)$ and the headings of trailers can be uniquely determined by setting $\dot \xi_k = 0$ for $k =4,5,6$, which admit the solutions
\begin{equation}\label{eq:CEC2}
\begin{aligned}
\xi_4(s) & =  - \arcsin(\frac{s d_1}{R}) \\
\xi_5(s) & = - \arcsin(\frac{sd_2}{Rc_1}) \\
\xi_6(s) & = - \arcsin(\frac{sd_3}{Rc_1c_2}),
\end{aligned}
\end{equation}
where $c_1 = \sqrt{1-(sd_1/R)^2}, c_2 = \sqrt{1-(sd_2/(Rc_1))^2}$.

By restricting the trailer to circular equilibrium configurations, we recover $\xi$ from $\bar X = (x,y,\theta_0,s)^\top \in \bar{\mathcal X} \subset \R^4$. Thus, planning searches over the 4D space $\bar{\mathcal X}$.

\section{Motion Primitive and Cost-to-go}

\subsection{Motion Primitive Generation}
\begin{figure}[t]
  \centering
  \includegraphics[width=0.4\textwidth]{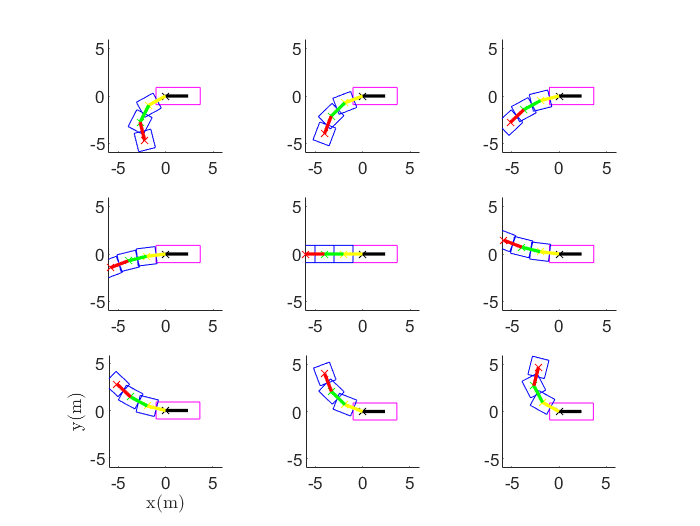}
  \caption{Initial states of the MPs (x- and y-axis units are meter).}
  \label{fig:X0pose}
\end{figure}

We sketch the motion primitive generation step for self-containess. Details can be found in \cite{LeuWanTom22}. MPs are generated by sampling initial-goal state pairs $(\bar{X}_0, \bar{X}_1)$ and solving the corresponding OCPs.
Define a compact set $\mathcal D = [-L_x,L_x]\times [-L_y,L_y] \times (-\pi,\pi] \times [-1,1]$, which is uniformly discretized into a finite set of states $\mathcal D_m = \{x_m\} \times \{y_m\} \times \{\theta_{0m}\} \times \{s_m\}$.
The established lattice-based methods, which apply MPs at $\bar X$ to reach other nodes, are not rotational-invariant: if one changes $\theta_0 \in \bar X$ and applies the same MP, the resultant new node usually no longer stays on lattice. They are however position-invariant in $xy$ directions, which means one only needs to generate MPs with initial states $\bar{X}_0 = (0, 0, \theta_0, s)$ where $\theta_0 \in \{\theta_{0m}\}$ and $s \in \{s_m\}$. The off-lattice method in \cite{wang2019improved,LeuWanTom22} is both position-invariant and rotation-invariant. Thus it only needs to generate MPs with $\bar{X}_0 = (0, 0, 0, s)$. 

An OCP for MP generation is solved for each pair $(\bar{X}_0, \bar{X}_1) \in \mathcal S_0 \times \mathcal S_1$, where $\mathcal S_0 = 0 \times 0 \times 0 \times \{s_m\}$ and $\mathcal S_1 = \mathcal D_m$.
Figure~\ref{fig:X0pose} visualizes all $\bar X_0$ when the interval $[-1,1]$ of $s$ is equally discretized into 9 points. From the symmetry of the states shown in Figure~\ref{fig:X0pose}, we only need to consider non-negative elements of $\{s_m\}$ which are denoted by $\{s_m^+\}$. Then, $\mathcal S_0 = 0 \times 0 \times 0 \times \{s_m^+\}$.
The OCP \eqref{eq:OCP} subject to (\ref{eq:fwd_car_dynamics2})-(\ref{eq:constraint_xu}) and boundary condition $(\bar{X}_0, \bar{X}_1)$, and can be solved using CasADi \cite{Andersson2019} and IPOPT~\cite{wachter2002interior}. 

The set of MPs is denoted as $P = \cup_{s \in \{s_m\}} P_s$, where $P_s$ is a subset of MPs whose initial state is $\bar{X}_0 = (0, 0, 0, s)$. The set $P$ is used by DE-AGT for motion planning. By applying the MPs, the tractor-trailer always moves from one circular equilibrium configuration to another. The system state evolves in $\mathcal X \subset \R^6$ during the transition between nodes.

\subsection{Modes of Motion Primitives}
\begin{figure}[t]
	\centering 
    \includegraphics[keepaspectratio=true, width=0.7\linewidth]{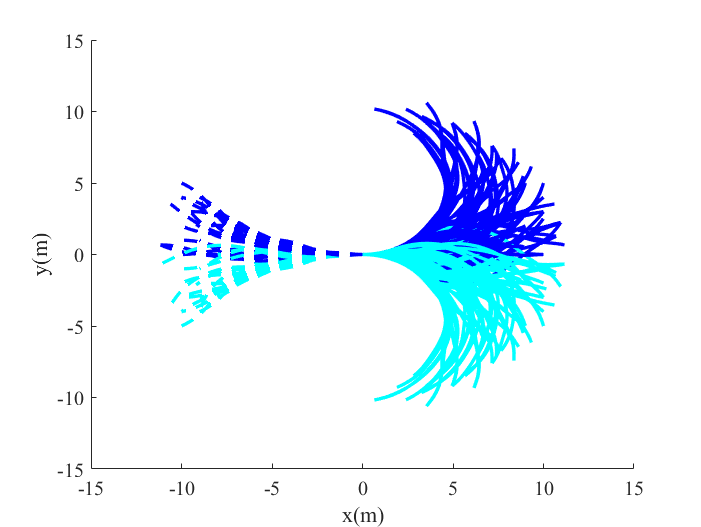}
	\caption{Modes of MPs. MPs in $P_s$ are classified into 4 modes based on their final states.}
	\label{fig:mode_illustration}
\end{figure}

We cluster the MPs in $P_s$ into several modes based on the 4 criteria: \emph{forward left}, \emph{forward right}, \emph{backward left}, and \emph{backward right}. The 4 criteria correspond to MPs with final states ending in the first, fourth, second, and third quadrant, respectively. The classes are also called the modes of MPs.
One example of the 4 modes for $\bar{X}_0 = (0, 0, 0, 0)$ is given in Figure~\ref{fig:mode_illustration}. 

\subsection{Cost-to-go Estimation} \label{Sec:Cost-to-go}
Another ingredient of DE-AGT is a heuristic function to estimate $J(\bar X, \bar X_g)$ which represents the cost-to-go from $\bar X$ to $\bar X_g$. Since we are interested in the cost-to-go from $\bar X \in \bar{\mathcal X}$ to a given $\bar X_g$, the cost-to-go can be abbreviated as $J(\bar X)$ - equivalently $J_{\bar X}$. Note that $J_{\bar X}$ is a valid formula because DE-AGT searches over $\bar {\mathcal X}$. We approximate $J_{\bar X}$ by an NN to be trained by using supervised learning. 

The first step is to generate the training data $(\bar X, J_{\bar X})$. Given $\bar X, \bar X_g$, the cost-to-go $J_{\bar X}$ can be obtained as the optimal cost by solving \eqref{eq:OCP}. To train an environment-indenpendent NN, we ignore the obstacle constraints in (\ref{eq:OCP}) to compute $J_{\bar X}$. Thus the cost-to-go is an underestimate of the true cost-to-go. We sample $n$ initial states $\bar{X}_s^{i}, i = 1, \cdots, n$ from set $\mathcal{D}_1 = [-L_x^{'}, L_x^{'}] \times [-L_y^{'}, L_y^{'}] \times (-\pi,\pi] \times [-1,1]$. Then $n$ OCPs are solved with boundary conditions $(\bar{X}_s^{i}, \bar{X}_g), i = 1, \cdots, n$,  providing $\bar n$ feasible trajectories. 
Hence we obtain $\bar n$ pairs $(\bar X^i_s,J_{\bar X^i_s})$ for $i = 1, \cdots, \bar n$ as training data, which are used to train a feed-forward NN using standard algorithms such as Bayesian Regression. The trained NN is a mapping from $\bar{X}_j$ to $J_{\bar{X}_j}$, and is denoted by $\mathsf{NNCost}: \bar{X}_j \rightarrow J_{\bar{X}_j}$.

\begin{rem}
    Usually, we need to train different NNs for different $\bar{X}_g$. However, for the trailer parking application, at the goal state $X_g$, the trailers are typically in line with the tractor. Also, using the position invariance and rotation invariance of the goal state, we only need the cost-to-go estimation for one goal state $\bar{X}_g = (0,0,0,0)$, \ie training one NN suffices our need.
\end{rem}

\begin{rem}
To avoid solving too many OCPs for training data, each feasible trajectory can be sampled for $\ell$ data points $(\bar{X}_j, J_{\bar X_j}), j = 1, \cdots, \ell$, where $J_{\bar X_j}$ approximates the cost-to-go from $\bar{X}_j$. By doing this for all $\bar n$ trajectories, we obtain $\bar n \times \ell$ data points. This treatment however compromises the training accuracy because the trajectory is not in $\bar {\mathcal X}$. 
\end{rem}

\section{The DE-AGT Algorithm} \label{sec:DE-AGT}
In this section, we describe the proposed DE-AGT algorithm in detail, focusing on two main features: online mode ranking enabled delayed expansion and integrating trajectory tracking with planning for improved goal-reaching.

\subsection{Mode Ranking}
The motivation for delayed expansion of MPs is illustrated in Figure \ref{fig:DelayedExpansionMotivation}.
With the MP classification and prioritization, DE-AGT expands the most promising mode of MPs first, which saves computation and finds a solution faster. DE-AGT will also expand the remaining modes when the same node is selected again for expansion.

One critical ingredient of delayed expansion is the online mode ranking, which is used to select the most promising MPs for expansion.
Once a node $n_c = (x_c, y_c, \theta_c, s_c)$ is selected for expansion, the MPs corresponding to $n_c$ from the set $P_{s_c}$ are ranked. 
For every $mp_i \in P_{s_c}$, the heuristic cost of expanding $mp_i$ at $n_c$ is computed according to 
\begin{equation}\label{mode_cost}
H(mp_i, n_c) =  \mathsf{cost}(mp_i) + \mathsf{NNCost}(n_{\mathrm{next}}),
\end{equation}
where $n_{\mathrm{next}}$ is the resulting node from applying $mp_i$ at $n_c$, $\mathsf{cost}(mp_i)$ is the cost of $mp_i$, and $\mathsf{NNCost}(n_{\mathrm{next}})$ is the predicted cost-to-go of $n_{\mathrm{next}}$.

Recall that MPs in $P_{s_c}$ have been divided into 4 modes. One can define the cost of a mode at $n_c$ as the average heuristic cost of the MPs in that mode. Finally, the modes are prioritized according to their costs. 

\subsection{Integrating Trajectory Tracking Control with Planning}
Exactly reaching a goal state is challenging. For state-lattice methods, reaching the goal state is only possible if the goal state is on-lattice. Otherwise, post-processing, e.g., trajectory optimization, is required. Similar difficulties apply to i-AGT. Without relying on a TPBVP solver that constantly performs goal connection checking, i-AGT terminates if it reaches a small neighborhood of the goal state. It is noteworthy that the tractor-trailer system has nonholonomy degree 2, implying that a small error might require quite a long maneuver to correct. Hence, for the tractor-trailer planning problem, the neighborhood size should be much smaller than a passenger car case. This renders the computational challenge of tractor-trailer planning particularly prominent.

\begin{figure}[t]
	\centering 
    \includegraphics[keepaspectratio=true, width=0.6\linewidth]{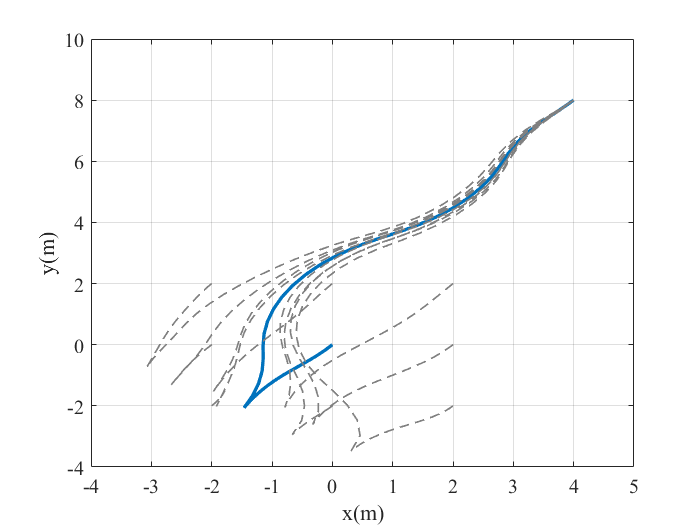}
	\caption{LQR trajectory tracking for the tractor-trailer. The solid blue line is the reference trajectory. The dashed lines are trajectories obtained by LQR control for different initial states.}
	\label{fig:LQRTracking}
\end{figure}

The DE-AGT algorithm overcomes this difficulty by integrating a light-weight trajectory tracking controller with the tree-based planning. 
The trajectory tracking controller is used to improve the chance that the specified neighborhood of the goal state can be reached. Figure \ref{fig:LQRTracking} shows an example of an LQR trajectory tracking controller for the tractor-trailer (\ref{eq:fwd_car_dynamics2}) to reduce the error. The LQR controller successfully reduces large initial state errors. 
The LQR controller is appealing for its computation efficiency and its robustness \cite{Khalil1996Robust}, but other options are also possible.
We repeatedly apply the LQR controller for goal connection whenever it is beneficial. 

\begin{figure}[t]
 \centering
   \begin{tabular}{@{}ccc@{}}
    \begin{minipage}{.23\textwidth}
    \includegraphics[keepaspectratio=true, width=1\linewidth]{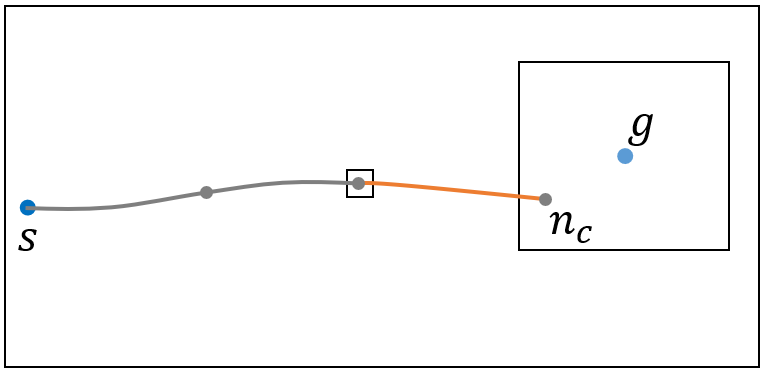}
    \captionof*{figure}{(a)} 
    \end{minipage} &
    \begin{minipage}{.23\textwidth}
    \includegraphics[keepaspectratio=true, width=1\linewidth]{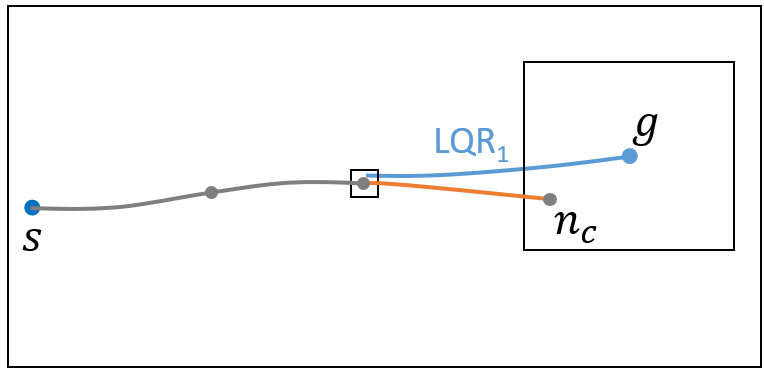}
     \captionof*{figure}{(b)}
   \end{minipage}
  \end{tabular}
  \begin{tabular}{@{}ccc@{}}
    \begin{minipage}{.23\textwidth}
    \includegraphics[keepaspectratio=true, width=1\linewidth]{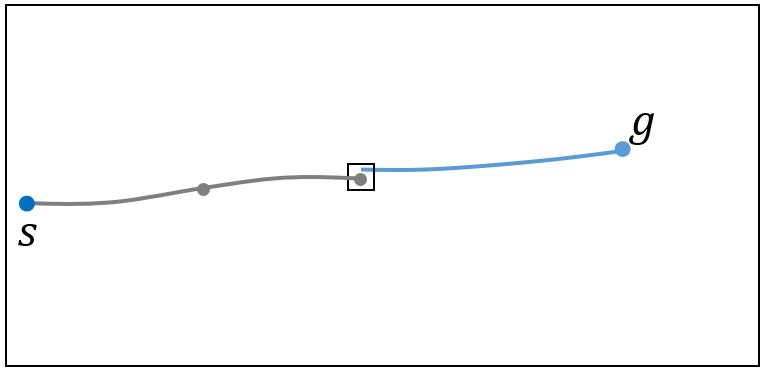}
    \captionof*{figure}{(c)} 
    \end{minipage} &
    \begin{minipage}{.23\textwidth}
    \includegraphics[keepaspectratio=true, width=1\linewidth]{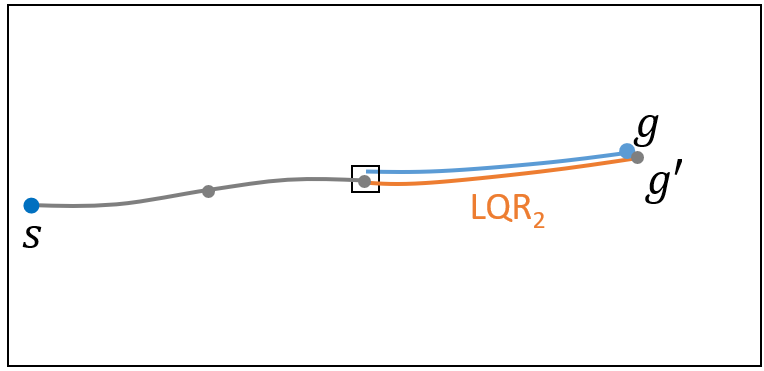}
     \captionof*{figure}{(d)}
   \end{minipage}
  \end{tabular}
    \caption{Integrating trajectory tracking with planning. (a) $n_c$ is in a neighborhood of $g$. The last segment of the trajectory (orange) is the reference trajectory for the first LQR. (b) LQR tracking controller is applied to generate a new trajectory (blue) to replace the orange trajectory. The new trajectory reduces the initial state error to a small error. (c) The blue trajectory is the reference trajectory for the second LQR. (d) The second LQR controller is applied to track the reference trajectory. If the resultant LQR trajectory (orange) reaches a small neighborhood of the goal and is collision-free, we accept this trajectory by adding the node $g'$ and the corresponding trajectory to the tree.}
  \label{fig:LQRTrackingIllustration}
\end{figure}

The method for integrating LQR trajectory tracking with planning is illustrated in Figure \ref{fig:LQRTrackingIllustration}. Whenever a newly added node $n_c$ is in a neighborhood (shown in Figure \ref{fig:LQRTrackingIllustration}(a) as the square box) of the goal state, we apply the LQR tracking controller to generate a new trajectory to replace the last segment of the trajectory that leads to $n_c$ (shown in Figure \ref{fig:LQRTrackingIllustration}(a) in orange).
The orange trajectory is the reference trajectory for the LQR controller. The trajectory generated using the LQR controller is shown as the blue line (LQR$_1$) in Figure \ref{fig:LQRTrackingIllustration}(b). Note that $n_c$ is the initial state of the reference trajectory and the LQR trajectory starts from $g$.
Here we used the property that for system (\ref{eq:fwd_car_dynamics2}), a trajectory from $g$ to $s$, $\overline{gs}$, can be generated using the trajectory from $s$ to $g$, $\overline{sg}$. The control inputs of $\overline{gs}$ are obtained by reversing the control inputs of $\overline{sg}$ and changing the sign of the velocities.
If the LQR trajectory reduces the error (indicated by the large square box and small square box in Figure \ref{fig:LQRTrackingIllustration}(b)) and is collision-free, the first part of the goal connection is successful.
In the second part, the LQR controller is applied again as shown in Figure \ref{fig:LQRTrackingIllustration}(d), where the blue line is the reference trajectory for the LQR control.
If the second LQR trajectory (orange line in Figure \ref{fig:LQRTrackingIllustration}(d)) reaches a small neighborhood of the goal and is collision-free, we accept this trajectory by adding the node $g'$ and the corresponding trajectory to the tree.

\subsection{DE-AGT}

\IncMargin{.5em}
\begin{algorithm}
\caption{DE-AGT}
\label{alg:DE-AGT}
$\mathcal{V} \leftarrow \{ x_\mathrm{init} \}$; $\mathcal{E} \leftarrow \emptyset$; $\mathcal{T} \leftarrow (\mathcal{V},\mathcal{E})$\;
$Q \leftarrow \{ x_\mathrm{init} \}$\;
$k \leftarrow 1$; $\mathrm{flag} \leftarrow \mathbf{False}$\;
\While{$k \le K$ $\mathbf{and}$ $\lnot \mathrm{flag}$} 
{
    $k \leftarrow k+1$\;
    $x_{\mathrm{select}} \leftarrow Q._{\mathrm{Top}}$\;
    \eIf {$d(x_\mathrm{select}, x_\mathrm{goal}) \le \epsilon_1$}
    {
        $\mathrm{flag} \leftarrow \mathbf{True}$\;
    }
    {   
            $(\mathcal{T}, Q) = \mathsf{ExpandTree}(\mathcal{T}, Q, x_{\mathrm{select}})$;
    }
}
\KwRet $\mathcal{T}, \mathrm{flag}$;
\end{algorithm}
\DecMargin{.5em}

\IncMargin{.5em}
\begin{algorithm}
\caption{ExpandTree}
\label{alg:ExpandTree}
$P_s \leftarrow \mathsf{FindMPs} (x_\mathrm{select})$\;
\If{$\mathsf{FirstTimeExpand}(x_{\mathrm{select}})$}
{
    $\mathrm{ModeCost} = \mathsf{ComputeModeCost}(x_{\mathrm{select}}, P_s)$\;
}
$\mathcal{M} = \mathsf{BestMode}(\mathrm{ModeCost})$\;
\ForEach{$mp \in \mathcal{M}$}
{
    $(x_{\mathrm{next}}, \tau) = \mathsf{Expand}(x_\mathrm{select}, mp)$\;
    \If{$\mathsf{CollisionFree(\tau})$}
    {
        \If{$d(x_\mathrm{next}, x_\mathrm{goal}) \le \epsilon_2$}
        {
            $(x_{\mathrm{next}}, \tau) = \mathsf{LQRConnect}(x_\mathrm{goal}, \tau)$\;
        }
        $\mathcal{V} \leftarrow \mathcal{V} \cup \{x_{\mathrm{next}}\}$; $\mathcal{E} \leftarrow \mathcal{E} \cup \tau$\;
        $Q._{\mathsf{Push}}(x_{\mathrm{next}})$\;
    }
}
$\mathrm{ModeCost} = \mathsf{UpdateModeCost}$\;
\If{$\mathsf{AllModeExpanded}(x_{\mathrm{select}})$} 
{
    $Q._{\mathsf{Delete}}(x_{\mathrm{select}})$\;
}
\KwRet $\mathcal{T}, Q$;
\end{algorithm}
\DecMargin{.5em}

The complete DE-AGT algorithm is given by Algorithm~\ref{alg:DE-AGT}.
DE-AGT maintains a search tree $\mathcal{T}$ and a priority queue $Q$.
The tree starts with the initial root node $x_{\mathrm{init}}$ and expands outwards by applying the available MPs until it reaches a neighborhood of the goal.
DE-AGT uses an A* heuristic to select the best node in $Q$ for expansion. Each node $x$ is assigned an $f$-value,
\begin{equation}\label{eq:f-value}
    f(x) =  g(x) + \alpha \mathsf{NNCost}(x),
\end{equation}
where $g(x)$ is the cost-to-come and $\alpha > 1$ is an inflation factor \cite{Likhachev2009Planning, wang2019improved}.
The operation $Q._{\mathrm{Top}}$ selects the best node in $Q$, i.e., that with the lowest $f$-value. Since the NN might output unreasonable cost-to-go estimate, we might use the following formula for $f(x)$:
$$
f(x) =  g(x) + \alpha \min(\max(\mathsf{RS}(x),\mathsf{NNCost}(x)),C),
$$
where $\mathsf{RS}(x)$ is the length of the Reeds-Shepp (RS) path for tractor only \cite{reeds1990optimal} and $C$ is a large positive constant. Given two configurations $(x_a, y_a,\theta_a)$ and $(x_b, y_b,\theta_b)$, one can compute the minimum length RS path between them.

In Algorithm~\ref{alg:ExpandTree}, $\mathsf{FindMPs}(x_\mathrm{select})$ finds the the set of MPs $P_s$ applicable at $x_{\mathrm{select}}$ according to the $s$ value of $x_{\mathrm{select}}$.
In lines 2-3, if $x_{\mathrm{select}}$ is selected for the first time, the cost of the MPs in $P_s$ are computed according to~(\ref{mode_cost}), and the cost of the modes of $P_s$ are computed and saved. In our case, the $\mathrm{ModeCost}$ is a vector with four entries since $P_s$ has 4 modes. If there are no MPs for a mode, this mode has infinite cost. 
$\mathsf{BestMode}(\mathrm{ModeCost})$ selects the mode with the lowest cost, and the MPs in $\mathcal{M}$ are used to expand $x_{\mathrm{select}}$ in line 5-11.
$\mathsf{Expand}(x_\mathrm{select}, mp)$ applies $mp$ to expand $x_{\mathrm{select}}$. The resulting trajectory is $\tau$ and the reached node is $x_{\mathrm{next}}$. 
If $x_{\mathrm{next}}$ is close enough to $x_{\mathrm{goal}}$ (line 8), the LQR controller is invoked for goal connection, which will potentially update $\tau$ and $x_{\mathrm{next}}$.
In line 10, the tree is updated by adding a new node $x_{\mathrm{next}}$ and new edge $\tau$.
$x_{\mathrm{next}}$ is pushed into queue $Q$.
After all MPs in $\mathcal{M}$ have been expanded, the mode cost of $\mathcal{M}$ is set to infinity (line 12) so that this mode will not be expanded. If all modes at $x_{\mathrm{select}}$ have been expanded, which also means that the costs of all modes are infinity, $x_{\mathrm{select}}$ is deleted from queue $Q$ (lines 13-14).

\begin{rem}
    Line 1 in Algorithm~\ref{alg:ExpandTree} implicitly imposes the constraint that the $s$ of $x_\mathrm{init}$ should belong to the set $\{s_m\}$.
\end{rem}

The goal connection method is given in Algorithm~\ref{alg:LQRConnect} as detailed in Section~\ref{sec:DE-AGT}.B and also illustrated in Figure \ref{fig:LQRTrackingIllustration}.
In line 1, the trajectory $\tau_{\mathrm{LQR}_1}$ corresponds to the blue trajectory in Figure \ref{fig:LQRTrackingIllustration}(b). In line 3, the trajectory $\tau_{\mathrm{LQR}_2}$ corresponds to the orange trajectory in Figure \ref{fig:LQRTrackingIllustration}(d).
$\mathsf{LQRConnect}$ uses an LQR trajectory tracking controller to update the trajectory $\tau$ such that the final state of the updated trajectory is much closer to the goal state.

\IncMargin{.5em}
\begin{algorithm}
\caption{LQRConnect}
\label{alg:LQRConnect}
$(x_{\mathrm{new}}, \tau_{\mathrm{LQR}_1}) = \mathsf{LQRTracking} (x_\mathrm{goal}, \tau)$\;
\If{$\mathsf{CollisionFree}(\tau_{\mathrm{LQR}_1})$ $\mathbf{and}$ $d(x_\mathrm{select}, x_\mathrm{new}) \le \epsilon_3$}
{
    $(x_{\mathrm{new}}, \tau_{\mathrm{LQR}_2}) = \mathsf{LQRTracking} (x_\mathrm{select}, \tau_{\mathrm{LQR}_1})$\;
    \If{$\mathsf{CollisionFree}(\tau_{\mathrm{LQR}_2})$ $\mathbf{and}$ $d(x_\mathrm{new}, x_\mathrm{goal}) \le \epsilon_4$}
    {
        $\tau \leftarrow \tau_{\mathrm{LQR}_2}$\;
        $x_{\mathrm{next}} \leftarrow x_{\mathrm{new}}$\;
    }
}
\KwRet $\tau, x_{\mathrm{next}}$;
\end{algorithm}
\DecMargin{.5em}

\section{Empirical Evaluation} \label{sec:sim}

\begin{figure*}[t]
 \centering
   \begin{tabular}{@{}ccccc@{}}
   \begin{minipage}{.19\textwidth}
    \includegraphics[width=\textwidth]{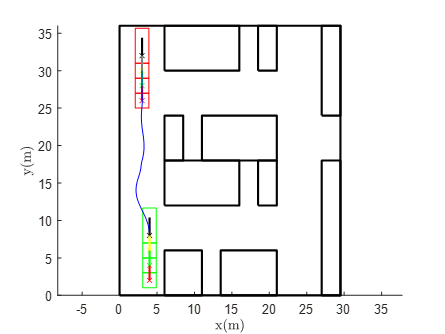}
     \captionof*{figure}{(a) Case 1.}
   \end{minipage} &
    \begin{minipage}{.19\textwidth}
    \includegraphics[width=\textwidth]{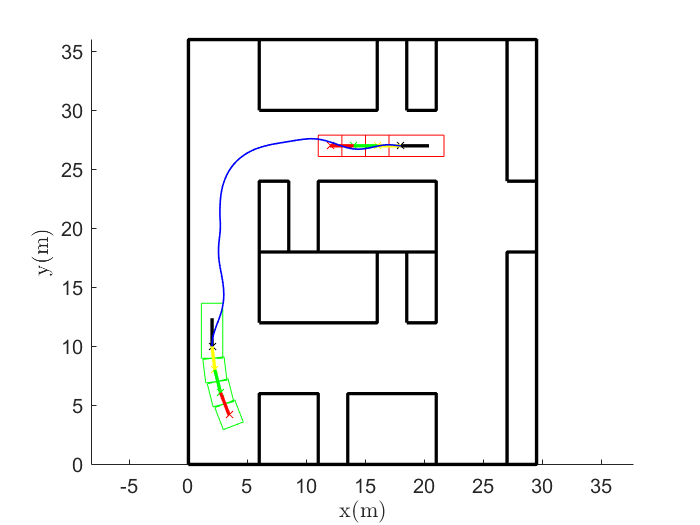}
    \captionof*{figure}{(b) Case 2.}
   \end{minipage} &
   \begin{minipage}{.19\textwidth}
    \includegraphics[width=\textwidth]{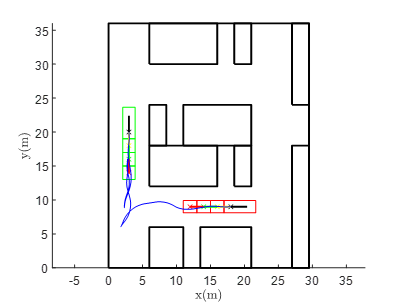}
    \captionof*{figure}{(c) Case 3.}
   \end{minipage} &
   \begin{minipage}{.19\textwidth}
    \includegraphics[width=\textwidth]{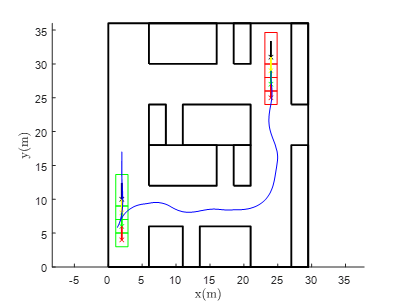}
    \captionof*{figure}{(d) Case 4.}
   \end{minipage} &
    \begin{minipage}{.19\textwidth}
    \includegraphics[width=\textwidth]{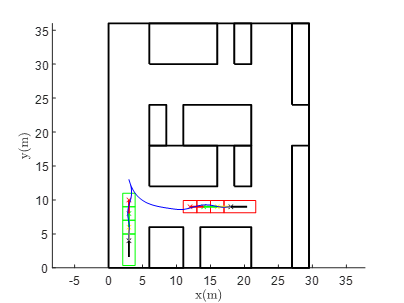}
    \captionof*{figure}{(e) Case 5.}
   \end{minipage}\\
   \begin{minipage}{.19\textwidth}
    \includegraphics[width=\textwidth]{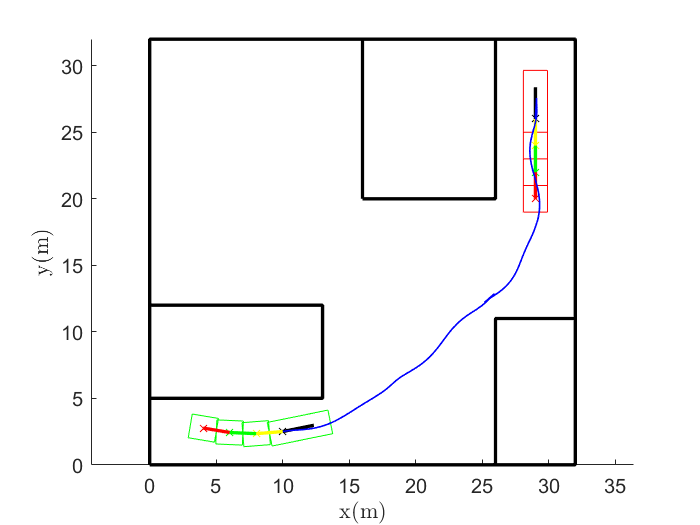}
    \captionof*{figure}{(f) Case 6.}
   \end{minipage} &
    \begin{minipage}{.19\textwidth}
    \includegraphics[width=\textwidth]{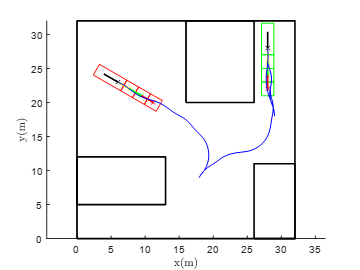}
    \captionof*{figure}{(g) Case 7.}
   \end{minipage} &
   \begin{minipage}{.19\textwidth}
    \includegraphics[width=\textwidth]{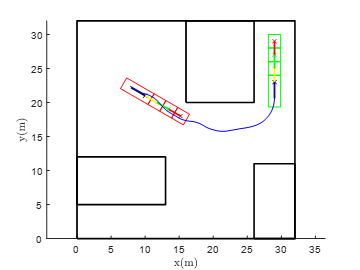}
    \captionof*{figure}{(f) Case 8.}
   \end{minipage} &
   \begin{minipage}{.19\textwidth}
    \includegraphics[width=\textwidth]{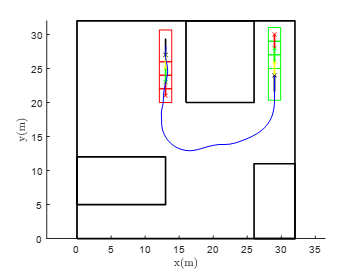}
    \captionof*{figure}{(h) Case 9.}
   \end{minipage} &
    \begin{minipage}{.19\textwidth}
    \includegraphics[width=\textwidth]{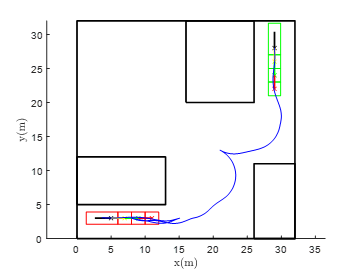}
    \captionof*{figure}{(i) Case 10.}
   \end{minipage}
  \end{tabular}
    \caption{Simulation results of the DE-AGT algorithm. Two test environments and 10 test cases. The initial position of the trailer is colored in green and the goal position is colored in red. The blue lines are the planned trajectory of the tractor.}
  \label{fig:sim}
\end{figure*}

\begin{table*}
\caption{Performance comparison of the DE-AGT algorithm and the i-AGT-RS algorithm}\label{tab:comparison}
\centering
\begin{tabular}{L{0.6cm}|L{1cm}|L{0.9cm}|L{1cm}|L{4.5cm}|L{1cm}|L{0.9cm}|L{1cm}|L{4.5cm}}
\toprule
\hline
\multirow{3}{0.6cm}{\centering Case No.} & \multicolumn{4}{c|}{DE-AGT}   & \multicolumn{4}{c}{i-AGT-RS} \\ \cline{2-9} & Planning Time [s] & \# of MP Explored & Path length [m] & Terminal State Error (abs) & Planning Time [s]  & \# of MP Explored & Path length [m]  & Terminal State Error (abs)\\ \hline
1  & \textbf{0.083}   & 89      & 24.48            & [0 0 0 0 0 0]  
   & 0.32             & 399     & 24.48            & [0 0 0 0 0 0] \\ 
2  & \textbf{0.13}    & 87      & 31.41            & \textbf{[0 0 0.001 0.002 0.005 0017]}  
   & 1.34             & 1281    & \textbf{29.21}   & [1.05 0.079 0.262 0.14 0.018 0.105] \\ 
3  & \textbf{1.12}    & 2142    & \textbf{48.31}   & \textbf{[0 0 0.004 0.008 0.0.014 0.026]}  
   & 15.26            & 25176   & 48.86            & [1.479 0.225 0.262 0.14 0.018 0.105] \\ 
4  & \textbf{18.98}   & 20893   & \textbf{62.05}   & [0.031 1.12 0 0 0 0]  
   & 213.27           & 348343  & 63.05            & [0.031 1.067 0 0 0 0] \\ 
5  & \textbf{0.42}    & 442     & 27.10            & \textbf{[0 0.001 0.007 0.012 0.023 0.043]} 
   & 3.10             & 3600    & \textbf{25.12}   & [1.172 0.221 0.262 0.14 0.018 0.105] \\
6  & \textbf{1.57}    & 3996    & 38.42            & \textbf{[0.004 0 0.032 0.061 0.064 0.061]}  
   & 3.76             & 7204    & \textbf{34.93}   & [0.069 0.158 0.062 0.183 0.306 0.429] \\ 
7  & \textbf{1.52}    & 4484    & \textbf{61.30}   & \textbf{[0 0 0.001 0.0.001 0.002 0.001]}  
   & 17.83            & 29206   & 75.52            & [0.792 0.542 0 0 0 0] \\ 
8  & \textbf{0.59}    & 515     & \textbf{29.89}   & \textbf{[0.42 0.38 0 0 0 0]}  
   & N/A*             & N/A     & N/A              & N/A \\
9  & \textbf{2.86}    & 22079   & \textbf{37.58}   & \textbf{[0.055 2.054 0 0 0 0]}  
   & N/A              & N/A     & N/A              & N/A \\
10 & \textbf{0.46}    & 839     & \textbf{64.33}   & [1.2 0.051 0 0 0 0]  
   & 12.22            & 14369   & 82.55            & [0.498 0.102 0 0 0 0] \\ \hline
\bottomrule
\end{tabular}
\begin{tablenotes}
\footnotesize
\item[*] *Solver failed or time out. The maximum computation time allowed is 500 seconds.
\end{tablenotes}
\end{table*}

In our simulations, the proposed DE-AGT and the baseline i-AGT-RS are applied to solve the motion planning for autonomous parking of a tractor-and-3-trailer system. For i-AGT-RS, the cost-to-go estimate is approximated by the length of the Reeds-Shepp path with tractor parameters. 
For fair comparisons, i-AGT-RS uses the same MPs as DE-AGT for planning. Compared to the baseline, DE-AGT uses a trained NN for cost-to-go estimation, delayed MP expansion to prioritize search, and integrates a trajectory tracking controller within the search.

We choose i-AGT-RS as the state-of-the-art is because ..., RRT sampling state missing the exact steering function and time-consuming for closed-loop tracking, sampling control space results in a high probability of rejection of nodes for backward maneuver, and thus it is almost impossible to find a feasible plan to a task requiring backward; A* with motion primitives for on-lattice search gives unfavorable maneuver, a path with low positioning accuracy, and long search time (for a large set of motion primitives and if the goal state is not on lattice). 

From the trailer model (2), the control domain is independent
from X. Accordingly, as in [13], [41], a motion
primitive for the 3-trailer system (2) can be defined as a
tuple of a normalized velocity and a constant steer angle
over certain period of time, and thus each X has exactly
the same set of motion primitives. i-AGT works as shown
in Fig. 2 where $X_0 = (3, 20, \pi/2, \pi/2, \pi/2, \pi/2)^\top,X_f =
(20, 27.5, 0, 0, 0, 0)^\top$; and it constructs a tree with 2617
nodes, and takes 7sec. It however undergoes heavy computation,
because the motion primitives explore the 6-dim
state space and lead to too many collision-free but undesirable
nodes. Rejecting nodes by exploiting the jack-knife
constraint ends up with 1219 nodes and 3.2sec, which is still
slow. This is because planning over the 6-dim state space is
dominated by the curse of dimensionality.

We tested our algorithm in two environments, shown in Figure \ref{fig:sim}(a) and \ref{fig:sim}(f), respectively. The simulation environments are designed to mimic a tractor-trailer moving in a large factory area where it needs to navigate through narrow aisles and park into narrow spaces. 
For each environment, we also tested our algorithm with different (start, goal) state pairs. 
The parameters for the tractor-trailer system are $L=2.396 \ \mathrm{m}$ and $d_1 =d_2=d_3=2 \ \mathrm{m}$. All simulations were conducted using Matlab R2021a on a 8-core Intel i7 3.7GHz desktop computer. Figure \ref{fig:sim} shows the planning results using DE-AGT for 10 different planning problems. 
Note that the heading of the tractor-trailer in the start and goal positions is important to differentiate these test cases. The tractor, shown as a longer rectangle, compared to the trailers, is located at the front.
The initial configuration of the tractor-trailer is shown in green color and the goal configuration is shown in red. 
In all tested cases, DE-AGT successfully finds the trajectory. 
 
The statistics of the planning results and comparison with i-AGT-RS are given in Table~\ref{tab:comparison}. 
The three key performance indices are planning time, path length, and terminal state error. From Table~\ref{tab:comparison}, it is seen that DE-AGT is faster to find the solution in all tested cases. Compared with i-AGT-RS, DE-AGT achieves, empirically, an average of 10x acceleration in terms of planning time.  
In terms of path length and terminal state error, DE-AGT also outperforms i-AGT-RS. DE-AGT finds shorter paths and has smaller terminal state errors in most cases. In cases 8 and 9 i-AGT-RS fails to find a solution within a 500 sec time limit.
The terminal state error is a 6D vector corresponding to the 6 states of the tractor-trailer. 
DE-AGT achieves small errors except for cases 4, 9, and 10. For cases 4 and 9, the errors along the y-axis are more than 1 m, and the error along the x-axis is more than 1 m for case 10. However, in those cases, the vehicle is aligned with the goal states. The vehicle is parallel to the y-axis (case 4, 9) or the x-axis (case 10), which means that we can simply drive the vehicle forward or backward and the error can be reduced to zero.

\section{Conclusion}\label{sec:conclusion}

This paper proposed the DE-AGT algorithm for kinodynamic planning of nonholonomic articulated vehicles with high-dimensional state space in cluttered environments.
DE-AGT inherits the efficiency of the state lattice-based methods by using pre-computed MPs that take care of the kinodynamic constraints. 
DE-AGT grows a trajectory tree using the pre-computed MPs and utilizes multiple heuristics to accelerate the search and find better motion plans. 
First, a data-driven method was employed to train a neural network for fast cost-to-go prediction.
Second, a delayed MP expansion method was developed, featuring MP classification and online mode ranking. 
DE-AGT expands the most promising mode of MPs first and expands the rest MPs only when necessary.
Third, a light-weight trajectory tracking controller was integrated into the planning process for constant goal connection to improve goal-reaching accuracy. Simulations using DE-AGT for tractor-trailer parking motion planning were shown. DE-AGT outperforms a competing approach (i-AG-RS) in terms of planning time, path length, terminal state error, and success rate. It is faster in all tested cases and achieves up to 26x acceleration compared to i-AGT-RS. Future work will focus on implementing DE-AGT on a full-size tractor-trailer system for autonomous parking. 

\balance

\bibliographystyle{IEEEtran}
\bibliography{PathPlanning}

\end{document}